\documentclass{article}

\usepackage{arxiv}

\usepackage[utf8]{inputenc} 
\usepackage[T1]{fontenc}    
\usepackage{hyperref}       
\usepackage{url}            
\usepackage{booktabs}       
\usepackage{amsfonts}       
\usepackage{nicefrac}       
\usepackage{microtype}      
\usepackage{lipsum}		
\usepackage{graphicx}
\usepackage{natbib}
\usepackage{doi}
\usepackage{fontawesome5}

\title{Biomedical reasoning in action:  Multi-agent System for Auditable Biomedical Evidence Synthesis}

\author{ {Oskar Wysocki}\thanks{Correspondence: oskar.wysocki@idiap.ch \href{https://youtu.be/Gxkpx_oPYbU}{\faIcon{youtube}~Short Video}} \\
	Idiap Research Institute\\
	Switzerland\\
	\And
	{Magdalena Wysocka} \\
	Idiap Research Institute\\
	Switzerland\\
    \And
    {Mauricio Jacobo} \\
    National Biomarker Centre (NBC)\\
    CRUK Manchester Institute\\
    United Kingdom\\
    \And
    {Harriet Unsworth} \\
    National Biomarker Centre (NBC)\\
    CRUK Manchester Institute\\
    United Kingdom\\
    \And
    {Andr\'e Freitas} \\
    NBC CRUK Manchester Institute, UK\\
    Idiap Research Institute, Switzerland\\
    Department of Computer Science\\
    University of Manchester, UK
}

\renewcommand{\shorttitle}{Biomedical reasoning in action}

\hypersetup{
pdftitle={Biomedical reasoning in action:  Multi-agent System for Auditable Biomedical Evidence Synthesis},
pdfsubject={q-bio.NC, q-bio.QM},
pdfauthor={David S.~Hippocampus, Elias D.~Striatum},
pdfkeywords={First keyword, Second keyword, More},
}

\begin{document}
\maketitle

\begin{abstract}
We present \textbf{M-Reason}, a demonstration system for transparent, agent-based reasoning and evidence integration in the biomedical domain, with a focus on cancer research. M-Reason leverages recent advances in large language models (LLMs) and modular agent orchestration to automate evidence retrieval, appraisal, and synthesis across diverse biomedical data sources. Each agent specializes in a specific evidence stream, enabling parallel processing and fine-grained analysis. The system emphasizes explainability, structured reporting, and user auditability, providing complete traceability from source evidence to final conclusions. We discuss critical tradeoffs between agent specialization, system complexity, and resource usage, as well as the integration of deterministic code for validation. An open, interactive user interface allows researchers to directly observe, explore and  evaluate the multi-agent workflow. Our evaluation demonstrates substantial gains in efficiency and output consistency, highlighting M-Reason’s potential as both a practical tool for evidence synthesis and a testbed for robust multi-agent LLM systems in scientific research, available at \url{https://m-reason.digitalecmt.com}.
\end{abstract}


\section{Introduction}
Large language models have evolved from static text generators to agentic ecosystems capable of planning, tool use, and collaboration. However, a key challenge remains: turning research prototypes into robust, user-facing systems where real-time interaction reveals both strengths and limitations. Public testbeds such as CAMEL’s “LLM society’’ \citep{li2023camel}, AutoGen’s multi-agent chat interface \citep{wu2023autogen}, LangChain mental-health chatbots \citep{singh2024revolutionizing}, and strategic-reasoning sandboxes \citep{sreedhar2024simulating} demonstrate how live demos expose issues missed by offline benchmarks, including API drift and unclear affordances. Surveys and industrial experiments further highlight the need for interactive, auditable platforms \citep{wang2024survey,ren2025towards,gao2025pharmagents}.

\begin{figure}[t]
    \centering
    \includegraphics[width=1\linewidth]{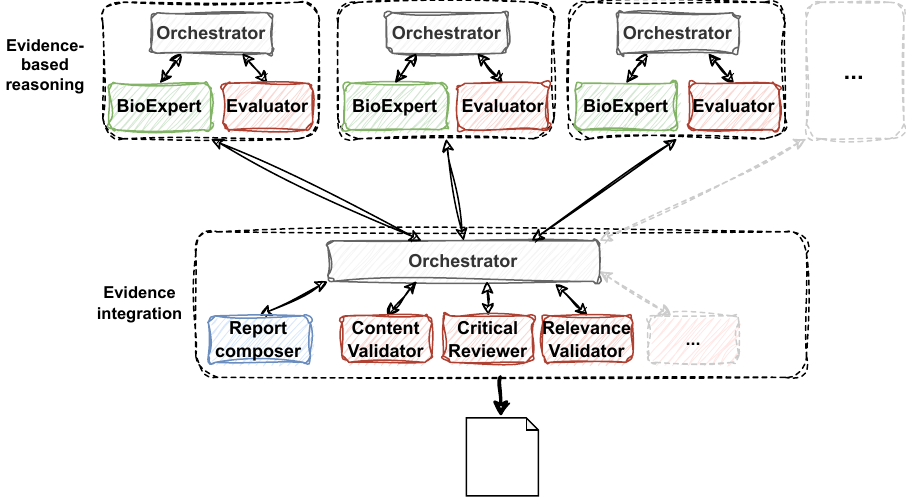}
    \caption{M-Reason system overview: visualization of agent orchestration, parallel processing of independent evidence modules, and extensible design enabling straightforward integration of new analytical agents.}
    \label{fig:universal_high_level}
\end{figure}

In biomedicine, demonstration frameworks such as CellAgent \citep{xiao2024cellagent}, BioInformatics Agent \citep{xin2024bioinformatics}, BioAgents \citep{mehandru2025bioagents}, and LORE \citep{10.1093/bib/bbaf070} help to lower barriers to workflow automation and analysis. Broad reviews and benchmarks \citep{gao2024empowering, mitchener2025bixbench,kehl2025use} map opportunities, while specialized systems tackle hypothesis generation \citep{kulkarni2025scientific}, claim analysis \citep{ortega2025sciclaims}, proteomics \citep{ding2024automating}, experimentation \citep{luo2024intention}, evidence synthesis \citep{wang2024accelerating}, oncology decision support \citep{lammert2024expert}, knowledge-graph diagnosis \citep{zuo2025kg4diagnosis}, and generalist medical models \citep{liu2025generalist}. Yet persistent gaps remain in guideline adherence, factuality, robustness \citep{mehandru2024evaluating,hager2024evaluation,wysocka2024large,hamed2025knowledge}, background-knowledge encoding \citep{wysocki2023transformers,gorski2025integrating}, domain-specific reasoning \citep{wysocka-etal-2025-syllobio}, and the need for reproducible, expert-auditable workflows \citep{wysocka2023systematic,pelletier2025evidence,wysocki2024llm}.

Against this backdrop, we present \textbf{M-Reason}: a system that puts recent advances in agent-based large language model (LLM) inference into practice for biomedical evidence synthesis. M-Reason directly responds to the community’s call for interactive, robust, and auditable reasoning platforms by organizing evidence retrieval, appraisal, and integration into a modular multi-agent workflow that is both transparent and extensible. In this framework, independent agents each analyze a distinct stream of biomedical evidence—such as clinical variants, pharmacogenomics, or enrichment statistics—before their findings are systematically synthesized into a unified, structured report.

By design, M-Reason tackles the persistent obstacles outlined above, including data heterogeneity, provenance tracking, and the challenge of transforming LLM prototypes into practical, user-facing systems, by prioritizing explainability and user engagement. Users can submit research questions and the analysis context in real time and observe the underlying reasoning process as it unfolds, with full visibility into each agent’s justifications and outputs at every stage. Through agent-level transparency, structured outputs, and continuous auditability, M-Reason serves as a testbed for evaluating and improving multi-agent LLM workflows in biomedicine, empowering researchers to perform evidence synthesis with the rigor, reliability, and adaptability required for high-impact scientific inquiry. 

The M-Reason platform is available at \url{https://m-reason.digitalecmt.com}, and on Github\footnote{\url{https://github.com/mjr-uom/CEE_DART_Navigator-main}}.

\begin{figure*}[t]
    \centering
    \includegraphics[width=.9\linewidth]{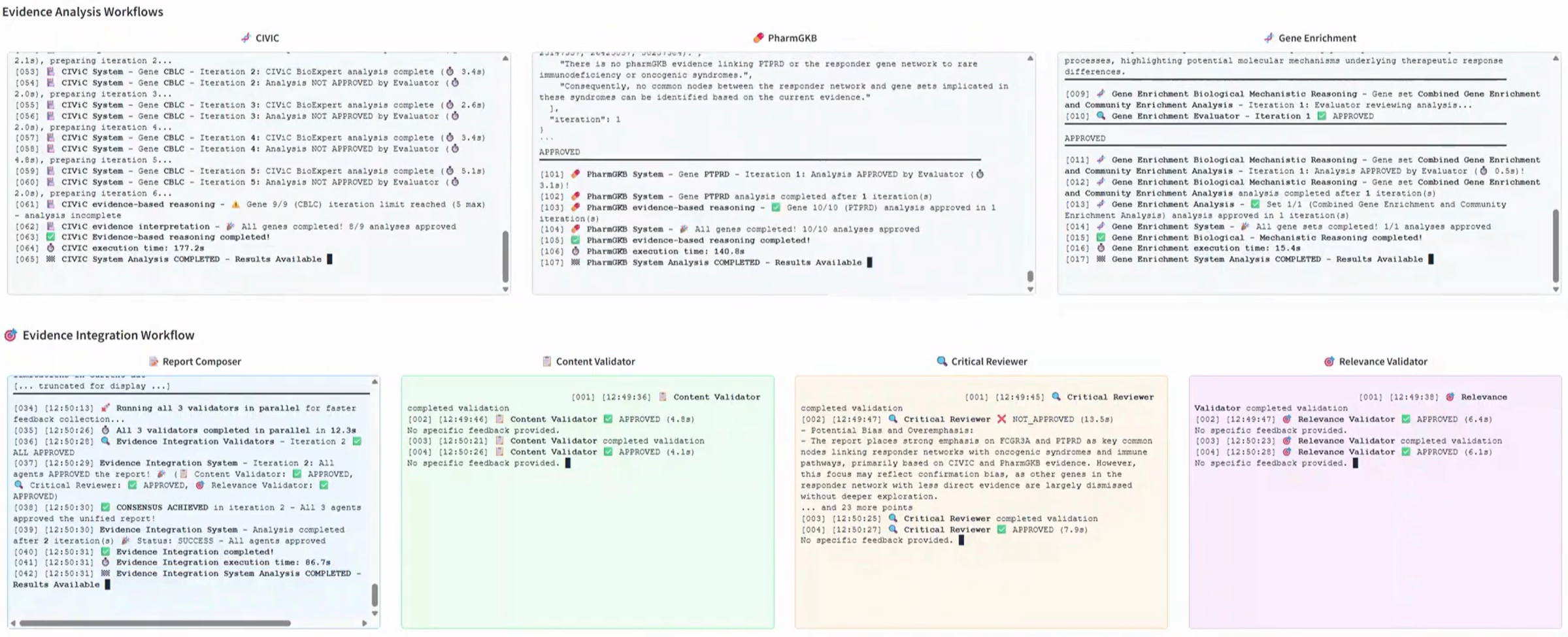}
    \caption{The M-Reason interface with seven message terminals, enabling users to observe the ongoing interpretation of evidence, structured report generation, and parallel feedback from validation agents.}
    \label{fig:ui_example}
\end{figure*}

\section{\textit{M-Reason} overview}

\textit{M-Reason} is organized around a modular architecture designed to streamline biomedical evidence retrieval, assessment, and synthesis. M-Reason is composed of several independent modules that operate in parallel to evaluate different types of evidence, followed by a synthesis module that integrates the results into a unified report. This architecture is intended to facilitate extensibility and maintain clear separation of responsibilities between components.

M-Reason is an integral part of the portal for numerical analysis and comparison of cancer patient samples, enabling direct transfer of analytical outputs to its agents and bridging the gap between bioinformatics pipelines and LLM-driven reasoning.

\subsection{Independent Evidence Retrieval and Assessment Modules}
Each evidence retrieval and assessment module in \textit{M-Reason} operates independently and is dedicated to a specific category of biomedical data. For example, one module focuses on clinical variant evidence from the CIVIC database\footnote{\url{https://civicdb.org/welcome}}\cite{griffith2017civic}, another handles pharmacogenomic information from PharmGKB\footnote{\url{https://www.pharmgkb.org/}}\cite{whirl2021evidence}, and a third is responsible for gene enrichment analysis using gProfiler\footnote{\url{https://biit.cs.ut.ee/gprofiler/gost}}\cite{10.1093/nar/gkad347}. In each module, evidence relevant to the user’s gene set and research question is first retrieved, then assessed by a specialized agent for clinical or biological significance. The independent nature of these modules allows them to run in parallel, supporting efficient processing and making it straightforward to incorporate additional knowledge sources or analytical approaches in the future.
Fundamentally, each agent serves as an intelligent filter, i.e. systematically reviewing the available evidence and selecting what is most relevant, important, or useful for the user’s specific question and context.

\subsection{Evidence Synthesis Module}
Once individual modules have completed their analyses, their outputs are collected by the synthesis module. This component is responsible for integrating the disparate streams of evidence into a coherent, comprehensive report. The synthesis process involves aggregating findings, resolving potential inconsistencies, and structuring the information in accordance with the user’s research objectives. To maintain consistency and quality, the synthesis module also coordinates an iterative review process involving validation agents who provide feedback on the draft report. The result is a single output that reflects all available evidence and addresses the original research context.

\section{\textit{M-Reason} Orchestration}

    
\subsection{Evidence analysis}

The evidence analysis process in the \textit{M-Reason} is organized as a system comprising three main agent roles: \textit{Orchestrator}, \textit{BioExpert}, and \textit{Evaluator}. This arrangement ensures systematic review of evidence and iterative quality control, with explicit mechanisms to prevent infinite feedback loops and maintain transparency at every stage.

\textbf{\textit{Orchestrator}} serves a coordinating function. It initiates the workflow by delegating tasks to the BioExpert and Evaluator agents but does not perform any reasoning or content analysis. The Orchestrator does not interact with any large language model; instead, its role is to sequence the workflow and return workflow status. All substantive analysis and evaluation are handled by downstream agents.

\textbf{\textit{BioExpert}} is responsible for analyzing the provided evidence in relation to the user's research context and question. The input prompt to the BioExpert is carefully constructed and adapts based on whether it is the first iteration or a subsequent revision:

$\bullet$ \textbf{First Iteration (Initial Analysis):} The agent receives a system prompt with detailed instructions and a user message that includes the research context, the user's question, and the relevant evidence. The prompt explicitly directs the BioExpert to analyze the evidence and answer the question in a structured format, providing relevance explanations, summaries, conclusions, and explicit citations to evidence sources.

$\bullet$ \textbf{Subsequent Iterations (Revision Mode):} For later rounds, the BioExpert receives a similar system prompt, but the user message additionally includes the previous analysis and detailed feedback from the Evaluator. The agent is instructed to revise its earlier output, addressing each point of feedback while preserving any valid content from prior iterations. This iterative prompt ensures the BioExpert focuses on continuous improvement rather than repeating errors.

\textbf{\textit{Evaluator}} receives a comprehensive review package after each BioExpert analysis. Its prompt consists of a system message detailing review criteria and a user message with the research context, question, evidence, and BioExpert's structured output. The Evaluator's response must begin with a binary decision---either ``APPROVED'' if the analysis meets quality standards, or ``NOT APPROVED'' followed by actionable, bulleted feedback. The evaluator assesses analysis on multiple criteria, including scientific accuracy, citation, clarity, and completeness.

The workflow allows for multiple iterations, up to a set maximum, to refine the analysis based on Evaluator feedback. If the Evaluator's response is ``NOT APPROVED,'' its feedback is routed back to the BioExpert for revision. The process continues until approval is achieved or the iteration limit is reached. Throughout, the Orchestrator simply delegates and tracks workflow status without influencing content, ensuring a clear separation of responsibilities and facilitating reproducibility. 

\subsection{Evidence Integration}

The Evidence Integration System in \textit{M-Reason} provides a higher-order synthesis layer that integrates outputs from the upstream evidence analysis pipelines: including CIViC, PharmGKB, and Gene Enrichment—and can be horizontally extended to incorporate additional sources. It adopts a multi-agent, consensus-based architecture, designed to ensure rigorous biomedical reporting through parallel review and specialized prompt engineering.

\subsubsection{Overview and Workflow}

Unlike the sequential BioExpert-Evaluator pattern used in individual evidence pipelines, the integration system coordinates a five-agents architecture with distinct roles: evidence consolidation (\textit{Orchestrator}), report composition (\textit{Report Composer}), and parallel multi-agent review (\textit{Content Validator}, \textit{Critical Reviewer}, \textit{Relevance Validator}). All review agents must reach unanimous consensus for report approval.

\textbf{\textit{Orchestrator}:} Responsible for workflow coordination. The Orchestrator parses the output files from each upstream analysis (CIViC, PharmGKB, Gene Enrichment), consolidates evidence into structured objects, and triggers the report synthesis process. This agent operates through Python logic alone, without LLM calls, and acts purely as a dispatcher.

\textbf{\textit{ReportComposer}:} Serves as the primary content creator. This LLM-based agent synthesizes evidence from all previous analysis systems into a structured report with four mandatory sections: \textit{potential novel biomarkers}, \textit{implications}, \textit{well-known interactions}, and \textit{conclusions}. Strict prompt constraints enforce i.a. the use of only provided evidence, bullet-point formatting, and evidence citation. The agent is also capable of revising its report based on combined feedback from all reviewers.

\textbf{\textit{ContentValidator}:} Focuses on structural integrity and content quality. It checks for the presence of all required sections, proper formatting, evidence-based statements, citations, and completeness of coverage across all evidence sources. It also ensures that no information outside of the provided evidence is introduced.

\textbf{\textit{CriticalReviewer}:} Provides adversarial analysis, focusing on bias detection, identification of unsupported claims, and suggesting alternative interpretations. This agent plays a key role in challenging assumptions and ensuring scientific rigor.

\textbf{\textit{RelevanceValidator}:} Ensures the report addresses the user's research question, with a particular focus on the classification of findings as novel or well-known, logical support for conclusions, and question alignment.

The review process is parallelized: the three reviewer agents independently assess each report version. Only unanimous approval by all reviewers permits the workflow to proceed to completion; otherwise, their collective feedback is sent to the \textit{Report Composer} for revision and resubmission.

\subsection{Prompt Engineering and Message Structures}

Each agent operates under a highly specialized system prompt, tailored for its function and for the current iteration. All agent prompts include explicit anti-hallucination instructions, requiring agents to refrain from introducing any information not directly present in the supplied evidence. 

\begin{figure}[b]
    \centering
    \includegraphics[width=1\linewidth]{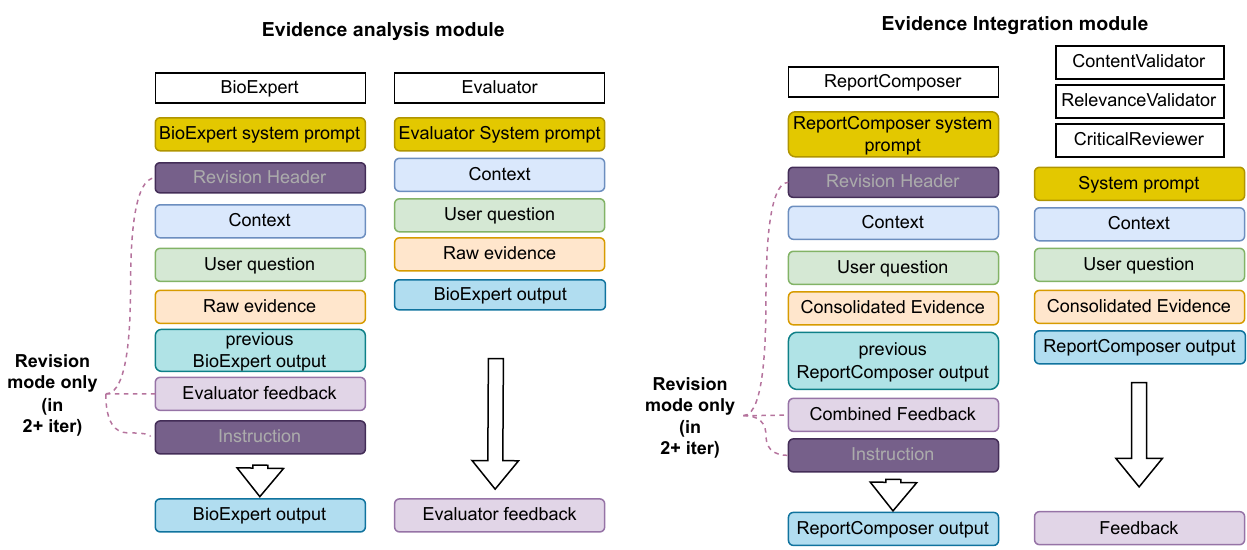}
    \caption{Prompt structure: Each agent receives a prompt that is context- and question-aware for every iteration. Prompts dynamically adapt to revision cycles by incorporating specific instructions to address reviewer feedback. Every prompt includes a detailed system message defining the agent’s role and responsibilities.}
    \label{fig:prompt_structure}
\end{figure}

The comprehensive structure of each input prompt to the LLM, as illustrated in Figure~\ref{fig:prompt_structure}, highlights the importance of including multiple components: the analysis context, user questions (to ensure tailored and specific reasoning), and the relevant evidence (raw or consolidated). Both the generation and evaluation agents require access to these inputs: the generator (\textit{BioExpert}) for analysis, and the \textit{Evaluator} for direct comparison and assessment of both evidence and analytic output. This design ensures precise instruction delivery, promotes reliable and context-aware analysis, and supports effective and transparent iterative evaluation (\textit{revision mode}).

\subsection{Design Principles and Rationale}

\textit{M-Reason} is developed around a set of core principles that address the main challenges of scalable, transparent, and context-sensitive biomedical evidence synthesis. These principles are consistently applied throughout the architecture, both in the Evidence Integration System and within every Orchestrator-BioExpert-Evaluator module, to ensure extensibility, scientific rigor, and utility for the end user.

\textbf{Novelty Detection and Classification:} A central focus across the entire \textit{M-Reason} is distinguishing between well-known and potentially novel findings. This is essential for biomedical researchers, as the value of the output depends on surfacing insights that go beyond established knowledge. By explicitly separating novel information from well-known facts, the system helps users identify discoveries with higher research impact.

\textbf{Extensible Multi-Source Evidence Integration:} \textit{M-Reason} is designed for seamless integration of multiple evidence sources. Its modular architecture allows for the addition of new knowledge bases simply by incorporating additional \textit{Orchestrator-BioExpert-Evaluator} modules, without requiring major changes to the rest of the pipeline. This ensures that the system remains adaptable and up-to-date as new biomedical databases become available.

\textbf{Structured Report Standardization:} All outputs, at every stage, are generated in structured JSON format with clearly defined sections. This reduces variability, enhances reliability, and makes it easier to pass information between modules. Structured reporting also facilitates downstream analyses, benchmarking, and integration with other tools.

\textbf{Parallel Processing and Unanimous Consensus:} Reviewer agents operate in parallel, and consensus-based approval is enforced throughout \textit{M-Reason}. This approach not only increases efficiency (multiple LLM-calls at the same time), but also ensures that multiple perspectives are incorporated, improving overall quality and trustworthiness.

\textbf{Source Citation and Traceability:} All synthesized outputs are fully traceable to their source evidence, with explicit citations (including direct links when available). This enhances transparency, user trust, and allows users to independently verify findings, directly addressing concerns about LLM hallucinations.

\textbf{User-Centric, Contextualized Inference:} At each stage, \textit{M-Reason} dynamically adapts its analysis and synthesis to the user’s specific research question and scientific context. Rather than merely aggregating or summarizing evidence, the system generates context-aware answers tailored to the precise query at hand. This capability is essential in biomedical research, where questions are highly nuanced and require interpretation within complex, domain-specific settings.

\textbf{Comprehensive Tracking of Inference Metrics:} Every module in \textit{M-Reason} logs token usage, runtime statistics, and inference metadata. This is vital for monitoring computational costs, especially with commercial APIs, and allows for objective comparison of system performance against manual curation.

By adhering to this structure, the system achieves extensibility, reliability, scientific validity, and practical value for biomedical researchers and domain experts.
This architecture establishes a robust foundation for reliable, extensible evidence synthesis, ensuring that unified reports address the user's research question with scientific rigor and transparency.

\section{\textit{M-Reason} Interface}

The user interface, as illustrated in Fig.\ref{fig:ui_example}, is designed to make the process of evidence-driven analysis both intuitive and transparent. At the outset, users can provide the overall context for their analysis and specify the research question they wish to investigate. To further tailor the analysis, users may manually input a list of genes, select gene lists generated by numerical analyses within the portal, or upload a JSON file containing their input data. This flexibility supports a wide range of user workflows, especially since M-Reason is integrated into a larger platform for deep learning results interpretation.

A core feature of the interface is its comprehensive evidence tracking. Throughout the analysis, users can monitor all pieces of evidence being considered and observe how they are processed by the system's agents. The interface is organized around seven distinct terminals, each dedicated to displaying the activities and communications of an individual agent. This layout allows users to follow the live, parallel execution and orchestration of the system in real time, gaining insights into the reasoning process at each step.

Upon completion of the analysis, the interface provides a final report summarizing the findings, which users can conveniently download as a PDF file. In addition, all logs, prompts, and intermediate outputs from each iteration are made available, supporting transparency and reproducibility. Execution metrics such as processing time, computational cost, token usage, the number of genes analyzed, and the number of iterations are also presented, giving users a comprehensive overview of the analysis and its resource requirements.


\begin{figure}[h!]
    \centering
    \includegraphics[width=1\linewidth]{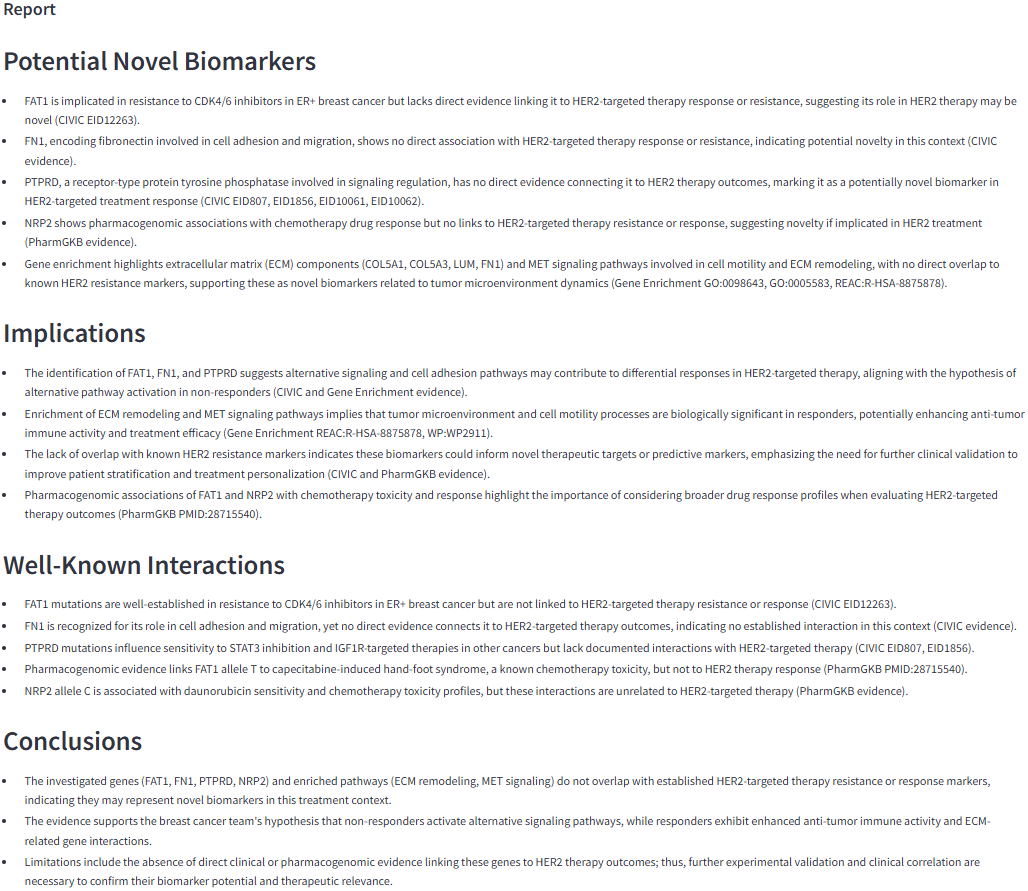}
    \caption{Exemplary structured report generated by M-Reason, summarizing integrated biomedical evidence and highlighting novel findings, implications, and source citations.}
    \label{fig:report_example}
\end{figure}

\section{\textit{M-Reason} Evaluation}

\begin{figure}[t]
    \centering
    \includegraphics[width=.6\linewidth]{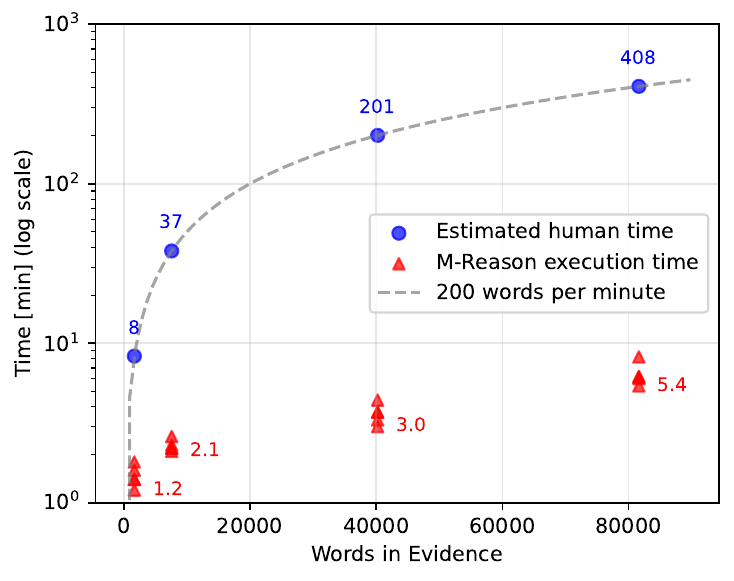}
    \caption{M-Reason report generation time compared to estimated human reading time for the same evidence sets. Y axis in log scale.}
    \label{fig:word_count_vs_reading_time}
\end{figure}

Our evaluation focused on two main aspects: quantitative time efficiency and qualitative output consistency. For the quantitative assessment, we measured the time savings offered by M-Reason in interpreting large volumes of evidence. We constructed four scenarios with increasing gene list sizes—$S_1$: 13, $S_2$: 28, $S_3$: 52, and $S_4$: 82 genes—corresponding to evidence sets ranging from 1,656 to 81,627 words. In each scenario, the gene list was expanded while retaining the genes from previous scenarios, allowing us to monitor whether significant findings were omitted as the input scale increased. The results demonstrate that M-Reason consistently preserved all critical findings, regardless of evidence size.

To assess consistency, we performed five independent executions for each scenario, maintaining the same context, question, and gene list. Both human experts and an LLM reviewed the outputs, confirming that the system reliably highlighted the same novel and well-known genes across runs.

For time efficiency, we baselined human expert reading speed at 200 words per minute, disregarding additional time for analysis or note-taking. As illustrated in Fig.\ref{fig:word_count_vs_reading_time}, M-Reason accelerated the reporting process dramatically. In the largest scenario ($S_4$), the system generated a comprehensive report approximately 135 times faster than a human would require to simply read the associated evidence.

While our evaluation is not exhaustive, it highlights M-Reason's capability and robustness in handling common reliability challenges found in LLM-based systems. Our results indicate that M-Reason offers both substantial time savings and consistent, trustworthy outputs when processing extensive and complex biomedical evidence.

\section{Conclusions}

We presented M-Reason, a modular, agent-based system for biomedical evidence integration and reasoning, focused on cancer research. M-Reason enables transparent, auditable workflows by organizing evidence retrieval, assessment, and synthesis into parallel, specialized agents. This approach improves reliability and user trust, while also exposing important tradeoffs between agent specialization, system complexity, and resource usage.

Our findings highlight that increasing agent specialization can enhance accuracy but comes with higher operational and maintenance costs. Similarly, combining LLM-driven analysis with deterministic code offers greater confidence, but at the expense of speed and flexibility—a tradeoff that remains central in system design.

By releasing M-Reason as an open, interactive demo, we empower users to directly engage with and critique multi-agent LLM workflows in biomedicine. We hope this system will foster further research into scalable, transparent, and robust AI for scientific evidence synthesis.

\section*{Limitations}

$\bullet$ The current evaluation of the system is based on only a small number of test cases, which may not fully capture its robustness or generalizability.

$\bullet$ Evidence integration is presently limited to three sources (CIViC, PharmGKB, and Gene Enrichment), though future versions are planned to incorporate additional sources such as PubMed for broader coverage.

$\bullet$ The system operates more as a deterministic workflow than a fully autonomous agent; while this improves predictability and reproducibility, it restricts adaptive decision-making capabilities.

$\bullet$ All evaluations have been conducted using the GPT-4.1-mini model, and system performance may vary with other or more powerful language models.

\section*{Acknowledgments}

This work was partially funded by the European Union’s Horizon 2020 research and innovation program (grant no.
965397) through the Cancer Core Europe DART project, and by the Manchester Experimental Cancer Medicine Centre and the NIHR Manchester Biomedical Research Centre.

\bibliographystyle{unsrtnat}
\bibliography{references}  






\end{document}